%% file: main.tex
\pdfoutput=1
\def\year{2022}\relax
\documentclass[letterpaper]{article} 
\usepackage{aaai22}  
\usepackage{times}  
\usepackage{helvet}  
\usepackage{courier}  
\usepackage[hyphens]{url}  
\usepackage{graphicx} 
\usepackage{natbib}  
\usepackage{caption} 
\DeclareCaptionStyle{ruled}{labelfont=normalfont,labelsep=colon,strut=off} 
\usepackage{algorithm}
\usepackage{algorithmic}
\usepackage{amsmath}
\numberwithin{equation}{section}

\usepackage{xcolor}
\numberwithin{equation}{section}

\usepackage{amssymb}
\usepackage{thm-restate}
\usepackage{subfig}

\usepackage{newfloat}
\usepackage{listings}
\floatstyle{ruled}
\newfloat{listing}{tb}{lst}{}
\floatname{listing}{Listing}
\title{Bayesian Optimization over Permutation Spaces}
\author{Aryan Deshwal,
        Syrine Belakaria, 
        Janardhan Rao Doppa, Dae Hyun Kim\\}

\affiliations{
School of EECS, Washington State University\\
{\{aryan.deshwal, syrine.belakaria, 
jana.doppa, daehyun.kim\}@wsu.edu}
}

\usepackage{bibentry}

\begin{document}

\maketitle

\begin{abstract}
Optimizing expensive to evaluate black-box functions over an input space consisting of all permutations of $d$ objects is an important problem with many real-world applications. For example, placement of functional blocks in hardware design to optimize performance via simulations. The overall goal is to minimize the number of function evaluations to find high-performing permutations. The key challenge in solving this problem using the Bayesian optimization (BO) framework is to trade-off the complexity of statistical model and tractability of acquisition function optimization. In this paper, we propose and evaluate two algorithms for {\em {\bf BO} over {\bf P}ermutation {\bf S}paces (BOPS)}. First, BOPS-T employs Gaussian process (GP) surrogate model with Kendall kernels and a {\bf T}ractable acquisition function optimization approach based on Thompson sampling to select the sequence of permutations for evaluation. Second, BOPS-H employs GP surrogate model with Mallow kernels and a {\bf H}euristic search approach to optimize expected improvement acquisition function. We theoretically analyze the performance of BOPS-T to show that their regret grows sub-linearly. Our experiments on multiple synthetic and real-world benchmarks show that both BOPS-T and BOPS-H perform better than the state-of-the-art BO algorithm for combinatorial spaces. To drive future research on this important problem, we make new resources and real-world benchmarks available to the community. 

\end{abstract}

\input{files/introduction.tex}

\input{files/problem_setup.tex}
\input{files/related_work.tex}
\input{files/technical_section.tex}

\input{files/experiments.tex}

\section{Conclusions}

We proposed and evaluated two effective Bayesian optimization algorithms with varying trade-offs for optimizing expensive black-box functions over the challenging input space of permutations. The results point to a key conclusion that it is important to use an appropriate model that exploits the specific structure of permutation spaces, which is different than the generic combinatorial space over categorical variables. We characterized the importance of this problem setting by describing three important real-world applications from the domain of computer-aided design of integrated circuits. Furthermore, we make all these benchmarks available to drive future research in this problem space. Future work includes studying extensions to handle high-dimensional spaces \cite{BOCK} and multiple objectives \cite{MESMO}.

\vspace{1.0ex}

\noindent {\bf Acknowledgments.} This research is supported in part by NSF grants IIS-1845922, OAC-1910213,and SII-2030159. 
\bibliography{main}
\clearpage
\input{files/appendix}

\end{document}

%% file: files/introduction.tex
\section{Introduction}
Optimizing expensive black-box functions is a common problem in many science and engineering applications \cite{bo_materialdesign,yang_protein_review,ACDesign,MOF}. An important class of such problems involve optimizing over the space of all permutations of a set of objects. For example, in the design of integrated circuits (ICs), we need to find placement of functional blocks to optimize performance via expensive simulations. Another example is to find a schedule for the jobs in additive manufacturing (parts can be built layer-by-layer) to optimize throughput via simulations. Bayesian optimization (BO) \cite{shahriari2016taking} has proven to be a successful approach for optimizing black-box functions in a sample-efficient manner. The key idea is to learn a surrogate statistical model, e.g., Gaussian process, and use it to select the sequence of inputs for function evaluation to uncover high-performing inputs. 

A large body of the BO literature focuses on continuous input spaces \cite{greenhill2020bayesian} with few recent works \cite{COMBO,BOCS,kdd_bo,IJCAI-2021} tackling the challenging setting of input spaces over discrete structures, e.g., sets, sequences, and graphs. Unlike continuous spaces, discrete spaces come with many unique challenges such as difficulty of defining a general representation, non-smoothness, etc. which require specialized treatment of different types of combinatorial structures \cite{sp_book}. In this paper, we consider the understudied space of all {\em permutations} of $d$ objects, which is equivalently characterized by the non-Abelian Symmetric group $\mathcal{S}_d$ of cardinality $d!$, consisting of all bijections from a set to itself. In contrast, all of the existing work on combinatorial BO is focused on the input space with {\em categorical/binary} variables which corresponds to the direct sums of abelian cyclic groups $\mathbb{Z}/c\mathbb{Z}$ of cardinality $c^d$, where each of the $d$ variables can take one of the $c$ categories. Combinatorial BO methods such as COMBO \cite{COMBO} cannot exploit the special characteristics of permutation spaces.  For example, they account for $d^d$ space $\gg$ $d!$ permutation space.

The key challenge to devise effective solutions for {\em {\bf BO} over {\bf P}ermutation {\bf S}paces (BOPS)} is to trade-off the complexity of statistical model and tractability of search to select permutations with high utility/acquisition values (e.g., expected improvement). We propose and evaluate two algorithms for BOPS by addressing this challenge. First, BOPS-T uses a simpler statistical model in the form of GP with Kendall kernels \cite{jiao2015kendall}. By considering the weight space view of the GP model and employing Thompson sampling as the acquisition strategy, we show that the acquisition function optimization is a Quadratic Assignment problem which can be solved using {\bf T}ractable Semi-definite programming relaxation based solvers. Second, BOPS-H uses a complex statistical model in the form of GP with Mallow kernels \cite{jiao2015kendall,mania2018kernel}. We employ expected improvement as the acquisition strategy and perform {\bf H}euristic search (local search with multiple restarts) to select the permutations for function evaluation. We analyze the theoretical properties of BOPS-T and in terms of regret bounds and show that it achieves a  sublinear (time) regret. Our comprehensive experiments on both synthetic and real-world benchmarks show that both BOPS-T and BOPS-H perform better than COMBO \cite{COMBO}.

\vspace{1.0ex}

\noindent {\bf Contributions.} The main contributions of this paper are:
\begin{itemize}
\setlength\itemsep{0em}
    \item Two BO algorithms for permutation spaces, namely, BOPS-T and BOPS-H, that make varying trade-offs between the complexity of statistical model and tractability of search for selecting permutations for evaluation.
    \item We theoretically analyze BOPS-T and show that it achieves sublinear (in time) regret bounds of $\mathcal{O}^{*}(d^{3/2}\sqrt{T})$, where $d$ and $T$ refers to dimensionality and time (no. of iterations) respectively and $\mathcal{O}^{*}$ denotes upto logarithmic factors. 
    \item We evaluate the efficacy of our algorithms and compare with the state-of-the-art COMBO algorithm on multiple synthetic and real-world benchmarks. 
    \item We make new resources and benchmarks based on important real-world problems available to the BO community to drive future research on this under-studied problem. The resources and source code of BOPS algorithms are available at \url{https://github.com/aryandeshwal/BOPS/}.
\end{itemize}

%% file: files/problem_setup.tex
\section{Problem Formulation}

In this paper, we consider optimization problems with the input space consisting of all permutations over $d$ objects. Given $[1, d] := \{1, 2, \cdots, d\}$, indexing the $d$ objects, a permutation is defined as a bijective mapping $\pi: [1, d] \mapsto [1,d]$. The set of  all permutations along with the composition binary operation $\left((\pi_1 \circ \pi_2)(x)= \pi_1 (\pi_2(x) \right)$ is known as the Symmetric group $\mathcal{S}_d$ which has a cardinality $|\mathcal{S}_d|$= $d!$. 

Let $f: \mathcal{S}_d \mapsto \mathbb{R}$ be a black-box objective function that is expensive to evaluate. Our goal is to optimize $f$ while {\em minimizing the number of function evaluations}: 
\begin{align}
    \pi^* = arg \min_{\pi \in  \mathcal{S}_d} f(\pi)
\end{align}

For a concrete example problem, consider the domain of design and optimization of integrated circuits (ICs). There are many applications in IC design, where we need to optimize over permutations of functional blocks of different granularity (small cells to processing cores). Some example objectives include performance and manufacturing cost. We need to perform expensive computational simulations to evaluate each candidate permutation.

We solve this problem using the Bayesian optimization (BO) framework. A BO algorithm is composed of two main components: 1) a {\em surrogate statistical model} built using past evaluations of the expensive black-box objective function; and 2) an {\em acquisition function} to capture the expected utility of evaluating a new input, which is optimized at each BO iteration to find the next best input for evaluation. Algorithm \ref{alg:bops} shows the generic BOPS approach. We provides concrete instantiations for these two components by handling the unique challenges of permutation spaces. In each BO iteration, we select the most promising permutation for evaluation (line 3) and update the statistical models using all the training examples (line 6). The best found permutation is returned at the end of maximum BO iterations (line 8).

\begin{algorithm}[H]
\caption{Bayesian Optimization over Permutation Spaces (BOPS)\label{alg:bops}}
\textbf{requires}: black-box objective $f(\pi)$, Gaussian process $GP(0, k)$ with Kernel over permutations $k$, Acquisition function $AF$
\begin{algorithmic}[1]
\STATE ${D}_0 \leftarrow$ small initial data; and $GP_0 \leftarrow GP(\mu_{D_0}, k_{D_0})$
\FOR{$j$=$1, 2, \dots$}
\STATE Acquisition function optimization to select the next permutation: $\pi_{j}$ = $arg \min_{\pi \in S_d} AF(\pi)$
\STATE Evaluate the selected permutation $\pi_{j}$ to get $f(\pi_j)$
\STATE Aggregate training data: ${D_j} \leftarrow D_{j-1} \cup \{\pi_j, f(\pi_j)\}$
\STATE Update GP posterior: $GP_j \leftarrow GP(\mu_{D_{j}}, k_{D_{j}}) $
\ENDFOR
\STATE \textbf{return} $\pi_{best}$ = $arg \min \{f(\pi_1), f(\pi_2) \cdots \}$
\end{algorithmic}
\end{algorithm}

%% file: files/related_work.tex
\section{Related Work}

Bayesian optimization (BO) methods for input spaces over discrete structures can be broadly classified into two categories based on the amount of available training data: {\em large} and {\em small} data settings. Our paper focuses on the small-data setting due to the unavailability of large training sets and the objective function evaluations being expensive.


\noindent {\bf BO methods for large-data setting.} Most of these methods employ the large set of training examples to learn embeddings of discrete structures in a latent space and perform continuous BO in the latent space \cite{reduction_continuous, reduction_continuous_2, bo_attribute_adjustment, grammar_vae}.  This setting is especially relevant for domains such as molecular design \cite{griffiths2017constrained,ts_bnn} and biological sequence design \cite{bio_sequence_design_1, bio_sequence_design_2}. 


\noindent {\bf BO methods for small-data setting.} Prior work in the small-data setting consider input spaces over binary/categorical structures: $x_1 \times x_2 \cdots \times x_d$ space, where each $x_i$ can take all possible values from a set of categories $\{0, 1, \cdots, c-1\}$, where $c$=2 and $c >2$ for binary and categorical variables respectively. This results in a search space of size $c^d$, which has a different structure than our considered setting of the space of all permutations of a set of $d$ objects. Prior methods in this setting can be classified based on their choice of surrogate models (linear models \cite{BOCS,kdd_bo}, random forests \cite{SMAC:TR2010}, Gaussian processes \cite{COMBO,LADDER}) or acquisition function optimization approaches (heuristic search \cite{SMAC:TR2010, COMBO, kdd_bo, LADDER}, principled mathematical optimization \cite{BOCS,PSR,MerCBO}), and combination thereof \cite{garrido2020dealing, deakin_discrete_bo, l2s_disco, amortized_bo_discrete_spaces}. 


\noindent {\bf Gaussian process models for discrete structures.}
COMBO \cite{COMBO} is the state-of-the-art method for small-data setting. It employs Gaussian process (GP) with diffusion kernels \cite{diffusion_kernel_original} defined over a combinatorial graph representation of the input space. However, COMBO's graph representation cannot exploit the special characteristics of permutation spaces. For example, for a space of $d$ objects, it requires $d$ subgraphs each with at least $d$ nodes to account for $d^d$ combinatorial space which is unnecessarily large for our usage. Another line of work considers GP models with kernels for specific structures including sets (double sum/embedding kernels  \cite{sets_bo, sets_kernels}), molecular graphs (optimal transport based kernel \cite{chembo}), and strings (string kernels \cite{string_bo,string_kernels}). We employ Kendall kernels and Mallow kernels \cite{jiao2015kendall,mania2018kernel} for permutation spaces which were shown to have superior performance on permutation based classification tasks.

%% file: files/technical_section.tex
\section{BO Algorithms for Permutation Spaces}

In this section, we provide two algorithms for BO over permutation spaces that make varying trade-offs between the complexity of statistical model and tractability of acquisition function optimization. First, BOPS-T employs a simple statistical model with an efficient Semi-definite programming (SDP) relaxation based optimization method. Second, BOPS-H employs a complex statistical model and performs heuristic search for optimizing the acquisition function. We employ Gaussian processes (GPs) \cite{GP-Book} as the surrogate model in both algorithms. GPs are effective statistical models commonly used for BO as they provide a principled framework for uncertainty estimation. They are fully characterized by a kernel $k$ \cite{kanagawa2018gaussian} which intuitively captures the similarity between two candidate inputs from the same input space.

\subsection{BOPS-T Algorithm}
\label{section:bopst}


\noindent {\bf Surrogate model.} The similarity between any two permutations ($\pi, \pi'$) can be naturally defined by considering the number of pairs of objects  ordered in the same way or in opposite ways. This is captured by the notion of the number of discordant pairs $n_d(\pi, \pi')$, also known as Kendall-tau distance \cite{kendall_tau_distance}. $n_d(\pi, \pi')$ counts the number of pairs of objects ordered oppositely by $\pi$ and $\pi'$ as defined below: 
\begin{align}
\begin{split}
    n_d(\pi, \pi') = \sum_{i<j} [1_{\pi(i) > \pi(j)} & 1_{\pi'(i) < \pi'(j)} \\
    &+ 1_{\pi(i) < \pi(j)} 1_{\pi'(i) > \pi'(j)}]
\end{split}
\end{align}

A related notion of concordant pairs $n_c(\pi, \pi')$ counts the number of object pairs ordered similarly by $\pi$ and $\pi'$: 
\begin{align}
 n_c(\pi, \pi')= \binom{d}{2} -  n_d(\pi, \pi') 
\end{align}

Kendall kernels \cite{jiao2015kendall} are positive-definite kernels defined over permutations using the notion of discordant and concordant pairs as follows:
\begin{align}
    k(\pi, \pi') =  \frac{n_c(\pi, \pi') -  n_d(\pi, \pi')}{\binom{d} {2}}  \label{eqn:kendall_kernel}
\end{align}
Because of their proven effectiveness over permutations \cite{jiao2015kendall,jiao2018weighted}, we propose using Kendall kernels with GPs as surrogate model in our BOPS-T algorithm. 

For our surrogate model, we consider the weight-space formulation of the GP. This weight-space formulation is essential for the SDP based acquisition function optimization approach described in the next section. In the {\em weight-space} view, we can reason about GPs as a weighted sum of basis functions $\phi$ =  $\{{\phi_i(\cdot)}\}$, i.e.,
\begin{align}
    w^T \phi(\cdot); \quad w \sim N(0, I) \label{weight_space}
\end{align}
where $N(\cdot)$ represents multi-variate Gaussian distribution and $I$ is the identity matrix.
Every kernel has a canonical feature map (as per the Moore-Aronszajn theorem \cite{moore_aronsazjn_theorem})  $\phi: \mathcal{S}_d \mapsto H_k$, $H_k$ being its associated Reproducing Kernel Hilbert Space (RKHS), that is employed as the basis function in \ref{weight_space}. The feature map expression for Kendall kernel (constructed by \cite{jiao2015kendall}) is given below:
\begin{align}
\phi(\pi) = \{\sqrt{\binom{d}{2}^{-1}} \left(1_{\pi(i) > \pi(j)} - 1_{\pi(i) < \pi(j)}\right) \}_{(1 \leq i < j \leq d)} \label{kendall_feature_space}
\end{align}


\noindent {\bf Acquisition function and optimizer.} In order to sequentially select the next permutation for evaluation guided by the learned surrogate model, we employ Thompson sampling as our acquisition function. Thompson sampling is a powerful, practitioner-friendly, and parameter-free approach for appropriately balancing the exploration vs. exploitation dilemma \cite{russo2014learning} in sequential bandit optimization. The key idea is to sample a function from the surrogate model's posterior and select its optimizer as the next permutation for evaluation. In the weight-space view of GPs, this corresponds to sampling a weight vector $\hat{w}$ from its posterior and solving the following optimization problem:
\begin{align}
    \pi_{next} = arg \min_{\pi \in \mathcal{S}_d} \hat{w}^T \phi(\pi) \label{eqn:TS_objective}
\end{align}
It should be noted that the sampled weight vector $\hat{w}$ is an exact function defined by GP (with Kendall kernel) over permutation spaces and has no approximation error when compared to the function space approach. This is in contrast to the common practice of using Thompson sampling over continuous spaces, where random Fourier features based weight-space representation of GPs is used which inevitably results in approximation error because of sampling a finite number of features from an infinite feature space.

We now show that the above acquisition function optimization problem (\ref{eqn:TS_objective}) is a Quadratic Assignment Problem (QAP) \cite{qap_reference}. To observe that, the objective in \ref{eqn:TS_objective} is written in an equivalent form in terms of $P_d$, the set of all possible permutation matrices $P$ of size $d \times d$, as follows:
\begin{align}
    \min_{P \in P_d} Tr(W  P A P^T) \label{eqn:QAP}
\end{align}
where $Tr$ is the matrix trace operation and $A$ is a $d \times d$ matrix defined as follows:
\begin{align*}
  A = \left\{
	\begin{array}{ll}
		1  & \mbox{if } i < j \\
		-1 & \mbox{if } i > j \\
		0 & \mbox{if } i == j
	\end{array}
\right.  \forall i, j \in [1, d]
\end{align*}
and $W$ is another $d \times d$ matrix given as follows:
\begin{align*}
  W = \left\{
	\begin{array}{ll}
		w_{\frac{(i-1)}{2} (2d-i) + (j - i)}  & \mbox{if } i < j \\
		0 & \mbox{if } i \geq j
	\end{array}
\right.  \forall i, j \in [1, d]
\end{align*}
Concretely, the equivalence of objectives in \ref{eqn:TS_objective} and \ref{eqn:QAP} can be seen as follows: 
\begin{align}
    Tr(W  P A P^T) = \sum_{i=1}^d  (WPAP^T)_{ii} \quad \label{eqn:trace_definition}
\end{align}
Equation \ref{eqn:trace_definition} is the definition of the trace of a matrix. Now, considering each entry $(WPAP^T)_{ii} $ in \ref{eqn:trace_definition}:
\begin{align}
    (WPAP^T)_{ii} &= \sum_{j=1}^d W_{ij} \cdot (PAP^T)_{ji} \\
    &= \sum_{j>i}^d w_{\frac{(i-1)}{2} (2d-i) + (j - i)} \cdot (PAP^T)_{ji} \label{eqn:w_definition}\\
    &= \sum_{j>i}^d w_{\frac{(i-1)}{2} (2d-i) + (j - i)}\cdot A_{\pi(j)\pi(i)} \label{eqn:permutation_definition}
\end{align}
where \ref{eqn:w_definition} follows from the definition of $W$ and \ref{eqn:permutation_definition} follows from the fact that pre-multiplying (post-multiplying) by a permutation matrix permutes the rows (columns) of $A$. Using \ref{eqn:permutation_definition} in \ref{eqn:trace_definition}:
\begin{align}
    Tr(W  P A P^T) = \sum_{i=1}^d \sum_{j>i}^d w_{\frac{(i-1)}{2} (2d-i) + (j - i)} \cdot A_{\pi(j)\pi(i)} 
\end{align}
By noting that $A_{\pi(j)\pi(i)}$ is exactly the feature map in \ref{kendall_feature_space} (upto multiplication by a constant $\sqrt{\binom{d}{2}^{-1}}$ which doesn't change the optimal solution), the equivalence between \ref{eqn:TS_objective} and  \ref{eqn:QAP} is established. 

Although, in general, Quadratic assignment probem is NP-hard \cite{qap_np_hard}, we leverage existing Semi-definite programming (SDP) based strong relaxations \cite{zhao1998semidefinite} to obtain good approximate solutions to the acquisition function optimization problem. Using the invariance of the trace under cyclic permutations and vectorization identity $(vec(APW) = (W^T \otimes A) vec(P))$, objective in \ref{eqn:QAP} is standardized as:
\begin{align}
    \min_{P, Q} &((W^T \otimes A) Q)  \label{eqn:vectorized_QAP} \\
    P &\in P_n  \nonumber\\
    Q &= vec(P) vec(P)^T  \nonumber
\end{align}
where $vec(P)$ is the column-wise vectorization of $P$. We leverage the clique-based SDP relaxation approach of \cite{sparse_sdp_qap} which can exploit matrix sparsity (e.g., zeros in the upper-triangular matrix $W$) for solving \ref{eqn:vectorized_QAP}. The key idea is to enforce semi-definiteness only over groups of $Q$'s entries (i.e., cliques) to get a relaxation that can be solved using fast and accurate algorithms. 

\subsection{BOPS-H Algorithm}


\noindent {\bf Surrogate model.} We propose to employ Mallows kernel which plays a role on the symmetric group $\mathcal{S}_d$ similar to the Gaussian (RBF) kernel on the Euclidean space. 
Given a pair of permutations $\pi$ and $\pi'$, the Mallows kernel is defined as the exponentiated negative of the number of discordant pairs $n_d(\pi, \pi')$ between $\pi$ and $\pi'$ i.e.
\begin{align}
    k_m {\pi, \pi'} = \exp (-l n_d(\pi, \pi')) 
\end{align}
where $l \geq 0$ is a hyper-parameter of the kernel similar to the length-scale hyper-parameter of the Gaussian kernels on Euclidean space. A key measure of the expressivity of a kernel is based on a property called {\em universality} which captures the notion of whether the RKHS of the kernel is rich enough to approximate any function on a given input space arbitrary well.
It was recently shown \cite{mania2018kernel} that Mallows kernel is {\em universal} over the space of permutations in contrast to the Kendall kernel discussed in the previous section. Therefore, Mallows kernels are more powerful than Kendall kernels and allows us to capture richer structure in permutations when used to learn GP based surrogate models. Indeed, our experiments also  demonstrate empirically the superior modeling capability of Mallows Kernel.


\noindent {\bf Acquisition function and optimizer.}
Unlike Kendall kernel, the feature space of Mallows Kernel is exponentially large \cite{mania2018kernel} making it practically inefficient to sample functions from the GP posterior (in the weight-space style as described earlier).Therefore, we propose to employ expected improvement (EI) as our acquisition function. The additional complexity of GP based statistical model with Mallows kernel makes the acquisition function optimization problem $\pi_{next}$ = $arg \min_{\pi \in \mathcal{S}_d} AF(\pi)$ is intractable for EI. Therfore, we propose to perform {\bf H}euristic search in the form of local search with multiple restarts that has been shown to be very effective in practice for solving combinatorial optimization problems. To search over only valid permutations $\pi \in \mathcal{S}_d$, at each local search step, we consider only those neighbors which are permutations of the current state. Otherwise, we will be searching over a huge combinatorial space with both valid (permutations) and invalid structures (non-permutations) as done by COMBO: may not result in producing a permutation from its acquisition function optimization procedure. Indeed, we observed this behavior in our experiments with COMBO. We use the modified local search procedure over permutations for both COMBO and our BOPS-H algorithm in experiments. 

\section{Theoretical Analysis for BOPS-T}
In this section, we analyze the theoretical properties of our BOPS-T algorithm in terms of regret metric \cite{srinivas2009gaussian}, which is a commonly used measure for analyzing BO algorithms. Note that there is no prior regret bound analysis for BO algorithms for EI even in continuous spaces. Hence, we leave the analysis of BOPS-H algorithm for future work. Let simple regret $R$ be defined as follows:
\begin{align}
    R = \sum_{t=1}^T (f(\pi_t) - f(\pi^{*}))
\end{align}
where $\pi_t$ is the permutation picked by the BO algorithm at time (iteration) $t$. In our case of using Thompson sampling as an acquisition function, it is natural to consider the expected form of this regret \cite{russo2014learning} where the expectation is taken over the distribution of functions as given by the GP prior with Kendall kernel. We analyze this expected form of regret, also known as Bayesian regret:
\begin{align}
    \mathcal{BR} = \sum_{t=1}^T \mathbb{E} (f(\pi_t) - f(\pi^{*}))
\end{align}
where the expectation is over the distribution of functions $f \sim GP(0, k)$. The below theorem bounds the Bayesian regret of our BOPS-T algorithm:

\begin{restatable}{theorem}{mt}
Let $f \sim GP(0, k)$ with Kendall kernel $k$ (\ref{eqn:kendall_kernel}), the Bayesian regret of the BOPS-T algorithm after $T$ observations $y_i$ = $f(\pi_i) + \epsilon_i, i \in \{1, 2, \cdots T\}$  with $\epsilon_i$ being Gaussian distributed i.i.d. noise  $\epsilon_i \sim N(0, \sigma^2)$ is :
$\mathcal{BR}$ = $\mathcal{O}^{*}(d^{3/2}\sqrt{T})$, 
where $\mathcal{O}^{*}$ denotes upto log factors.
\end{restatable}
{\noindent \bf Proof.} The key quantity in bounding the regret of Bayesian optimization with Gaussian processes (also known as GP bandits) is an information-theoretic quantity called as {\em maximum information gain} $\gamma_T$ \cite{srinivas2009gaussian} that depends on the kernel $k$ and intuitively captures the maximum information that can be gained about $f$ after $T$ observations, i.e., 
\begin{align}
    \gamma_T = \max_{A \subset \mathcal{S}_d, |A| = T} I(\mathbf{y_A}; f)
\end{align}
where $I$ is the mutual information and $A$ is a subset of permutations with corresponding function evaluations $\mathbf{y_A}$.

\cite{russo2014learning} proved the Bayesian regret for Thompson sampling by characterizing it in terms of upper confidence bound based results from \cite{srinivas2009gaussian}:
\begin{restatable}{proposition}{dl}
(Proposition 5 \cite{russo2014learning}). If $|X| < \infty$, $\{f(x): x \in X\}$ follows a multivariate Gaussian distribution with marginal variances bounded by 1, the Bayesian regret for Thompson sampling based bandit policy is given as: 
\begin{align}
     \mathcal{BR} = 1 + 2 \sqrt{T\gamma_T \ln{(1 + \sigma^2)^{-1}} \ln{\left(\frac{(T^2+1) |X|}{\sqrt{2\pi}}\right)}} \label{prop:russo}
\end{align}
where $X$ is the action space.
\end{restatable}

This proposition is directly applicable in our setting because the action space, being the cardinality of the symmetric group $\mathcal{S}_d$, is finite (i.e., $|\mathcal{S}_d|$ = $d!$) and the function $\{f(\pi): \pi \in \mathcal{S}_d\}$ follows a multivariate Gaussian distribution (by definition of Gaussian process with Kendall kernel). We compute the specific terms in the right-hand side of \ref{prop:russo} that are applicable in our setting to prove the regret bound. 

The maximum information gain for kernels with finite feature maps can be computed in the weight-space form (Sec \ref{section:bopst}) as a special case of linear kernel \cite{srinivas2009gaussian}.
\begin{align}
    \gamma_T  \leq C \log |I + \sigma^{-2}K|
\end{align}
where $C$ = $1/2 \cdot (1-1/e)^{-1}$ is a constant, $K$ is a $T\times T$ matrix with each entry $K_{ij}$ = $k(\pi_i, \pi_j)$. As per kernel trick,
\begin{align}
    K = \Phi^T \Phi
\end{align}
where $\Phi$ is a matrix with $\Sigma^{1/2} \phi(\pi_i), i \in \{1, 2, \cdots, T\}$ as the columns (\ref{kendall_feature_space}). Therefore,
\begin{align}
    \gamma_T  \leq C \ln |I + \sigma^{-2} \Phi^T \Phi|
\end{align}
By Schur's complement:
\begin{align}
    \gamma_T  \leq C \ln |I + \sigma^{-2} \Phi^T \Phi| \leq C \ln |I + \sigma^{-2} \Phi \Phi^T|
\end{align}
By Hadamard's inequality:
\begin{align}
    \gamma_T  &\leq C \ln |I + \sigma^{-2} \Phi \Phi^T| \\
     &\leq C \sum_{i=1}^{\binom{d}{2}} \ln(1 + \sigma^{-2} \lambda_i)
\end{align}
where $\{\lambda_1, \lambda_2, \cdots \}$ is the eigenvalue set of the matrix $\Phi \Phi^T$.

By Gershgorin circle theorem \cite{varga2010gervsgorin}, all the eigenvalues of a matrix is upper bounded by the maximum absolute sum of rows, i.e. $\lambda_i \leq d^2 T$ with the assumption that $\| \Sigma^{1/2}\phi(\pi) \| \leq 1$.
\begin{align}
    \gamma_T = O(d^2 \ln (d^2T)) \label{gamma_final}
\end{align}

Now, using Stirling's approximation, we can bound the $\ln(|X|)$ term in \ref{prop:russo}, where $|X| = |\mathcal{S}_d|$ in our case:
\begin{align}
    \ln(|\mathcal{S}_d|) = O(d \ln d) \label{sterling}
\end{align}

Plugging \ref{gamma_final} and \ref{sterling} in \ref{prop:russo} and ignoring constants, we get the following expression:
\begin{align}
         \mathcal{BR} &= O(\sqrt{T d^2 \ln d^2T (\ln T^2 + d \ln d)}) \\
       \mathcal{BR}  &= O(\sqrt{ ( T d^2 \ln d^2T \ln T^2 +  T d^3 \ln d^2T  \ln d)}) \\
       \mathcal{BR}  &= \mathcal{O}^{*} (d^{3/2} \sqrt{T})
\end{align}

Hence, ignoring log factors, our proposed BOPS-T algorithm achieves sublinear (time) regret.

%% file: files/experiments.tex
\section{Experiments and Results}
\begin{figure*}[h!]
\centering
\subfloat{
\includegraphics[width=0.30\textwidth]{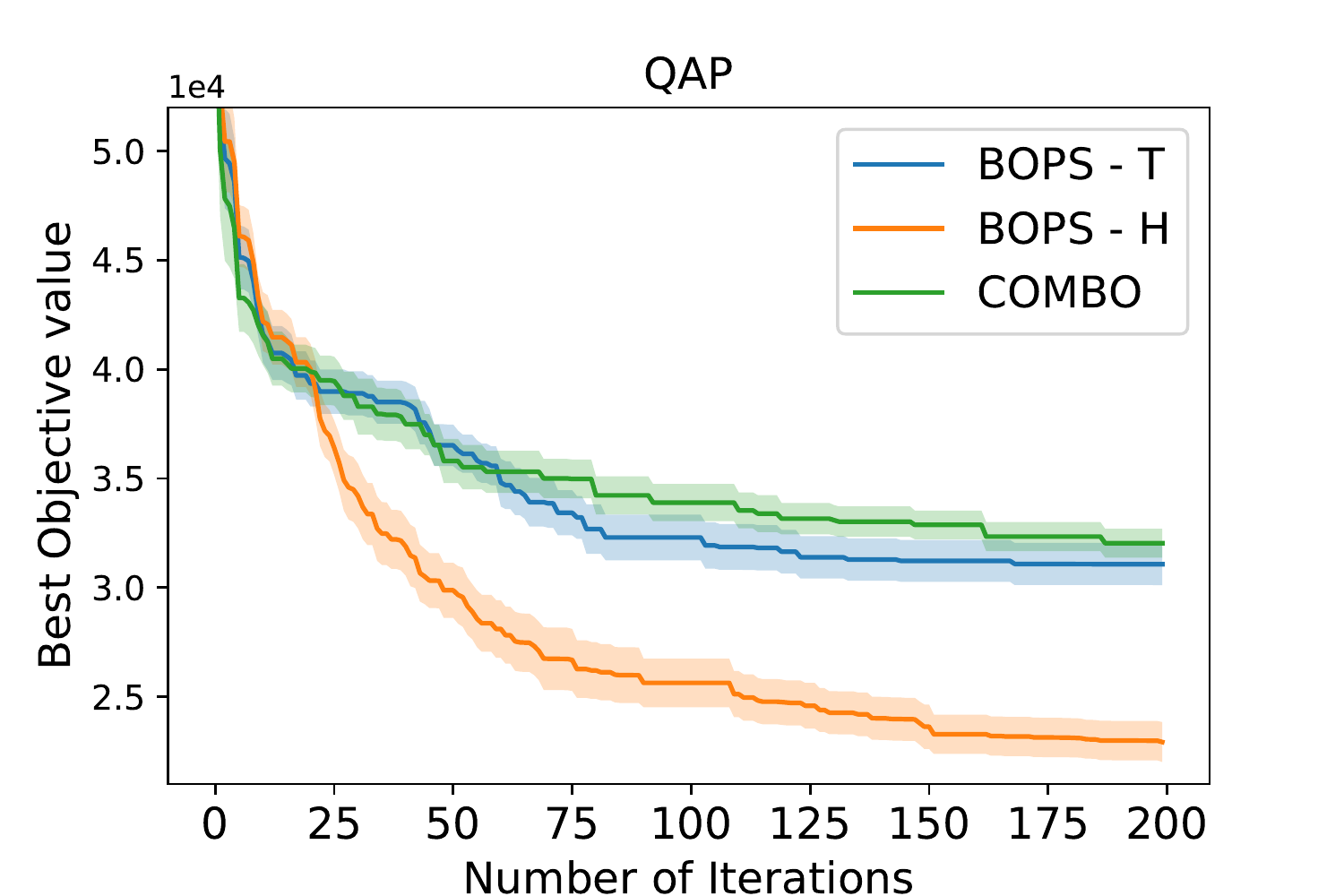}
\label{fig:qap}
}
\subfloat{
\includegraphics[width=0.30\textwidth]{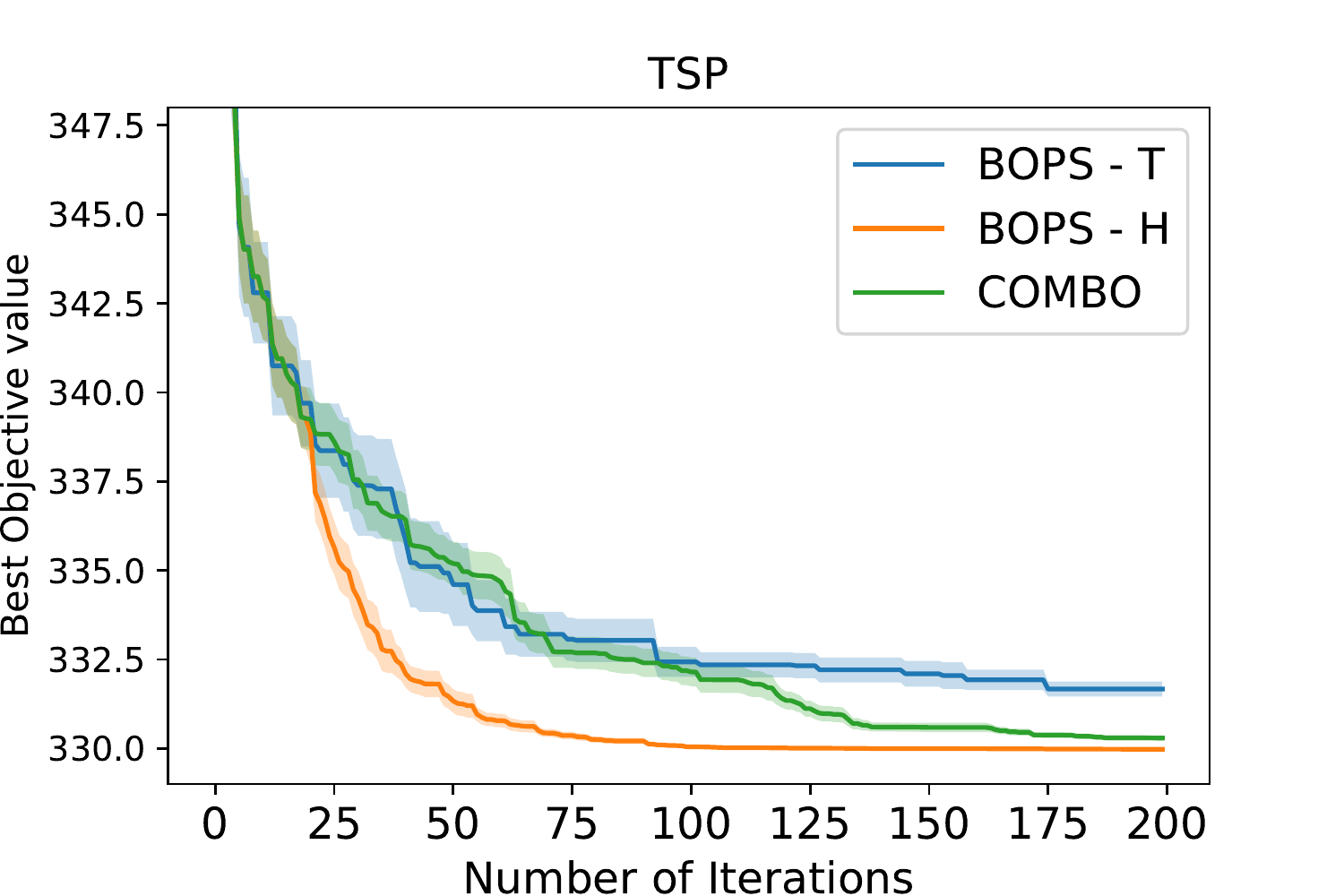}
\label{fig:tsp}
}
\subfloat{
\includegraphics[width=0.30\textwidth]{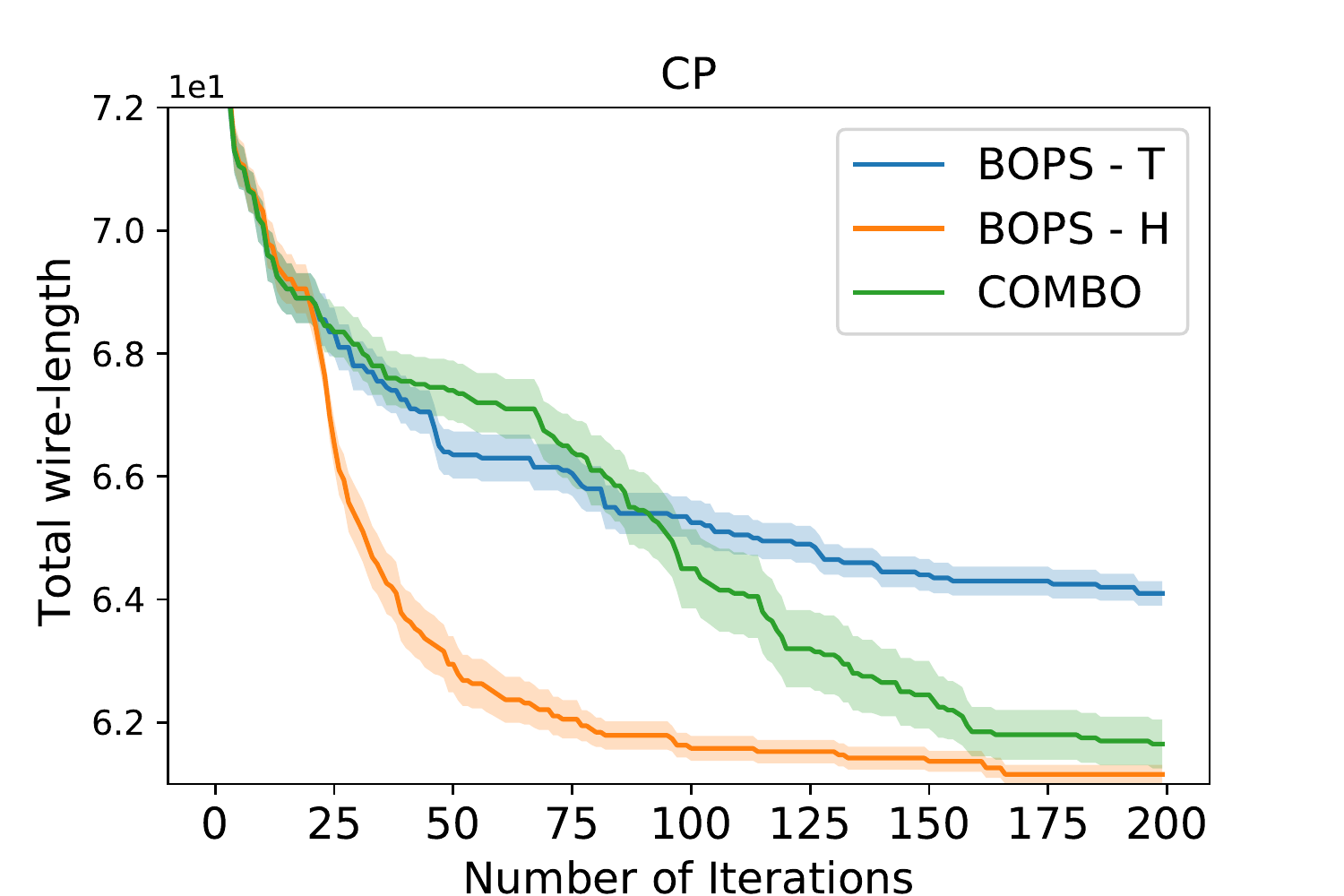}
\label{fig:pl}
} 
\quad
\subfloat{
\includegraphics[width=0.30\textwidth]{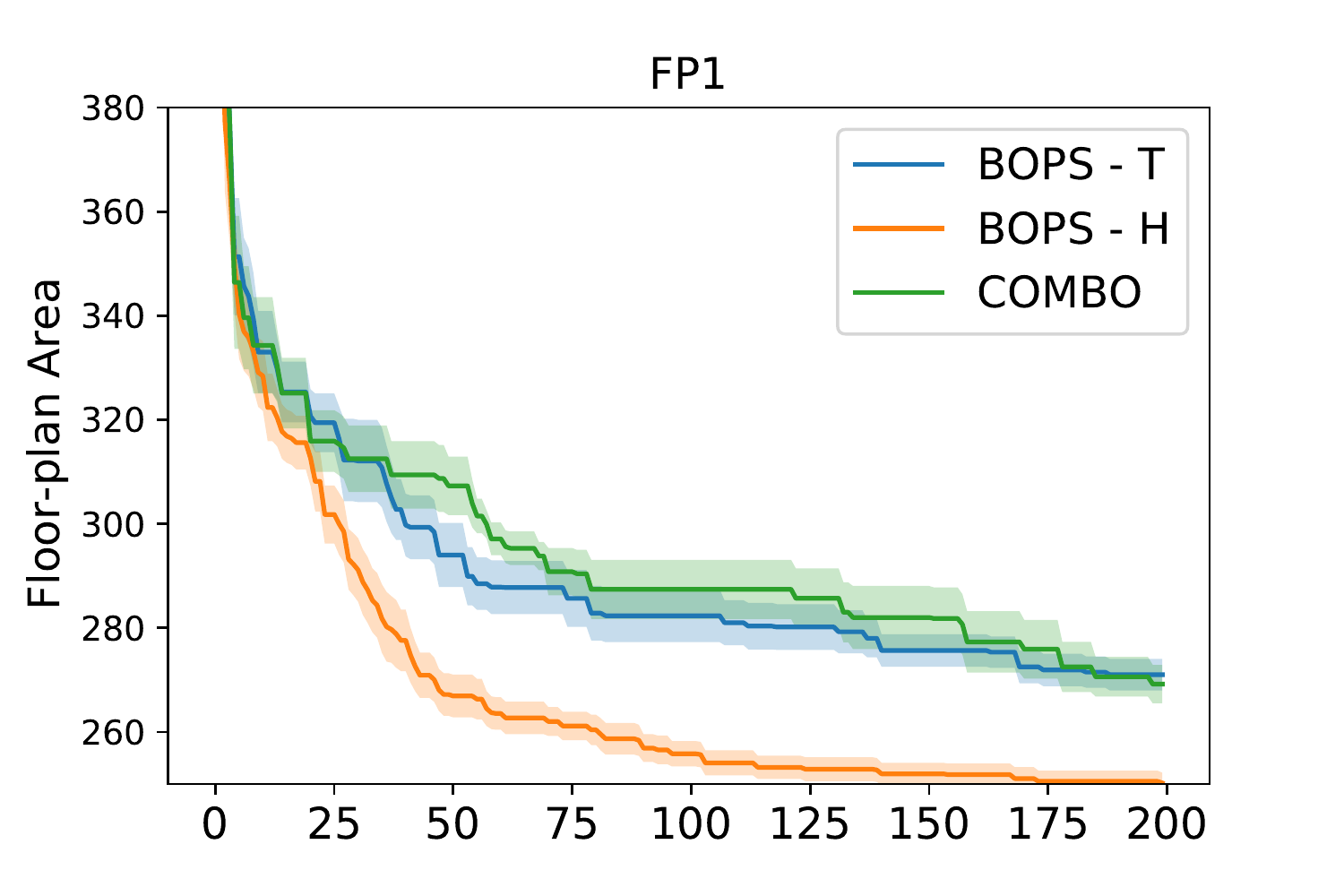}
\label{fig:fp_1}
}
\subfloat{
\includegraphics[width=0.30\textwidth]{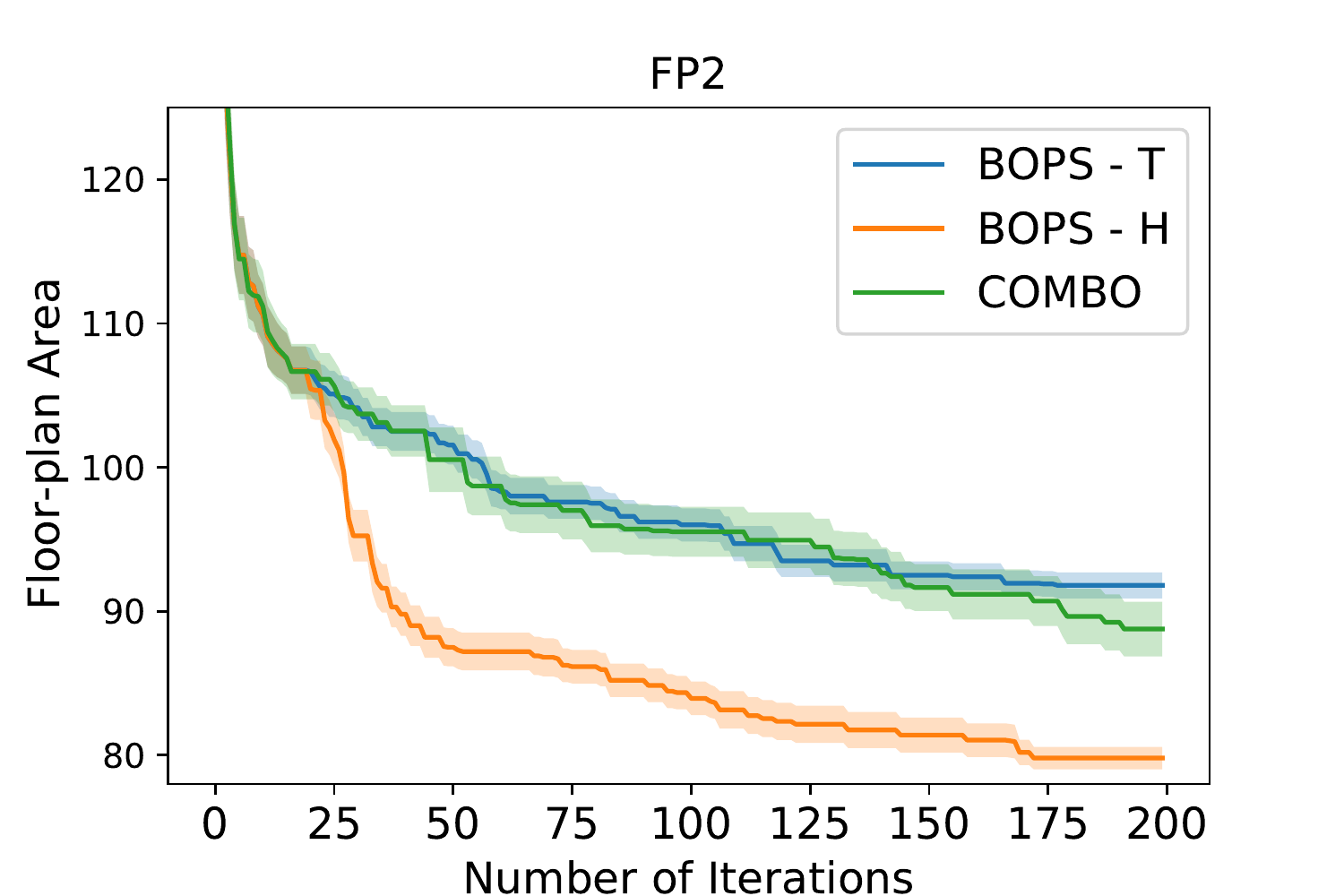}
\label{fig:fp_2}
}
\subfloat{
\includegraphics[width=0.30\textwidth]{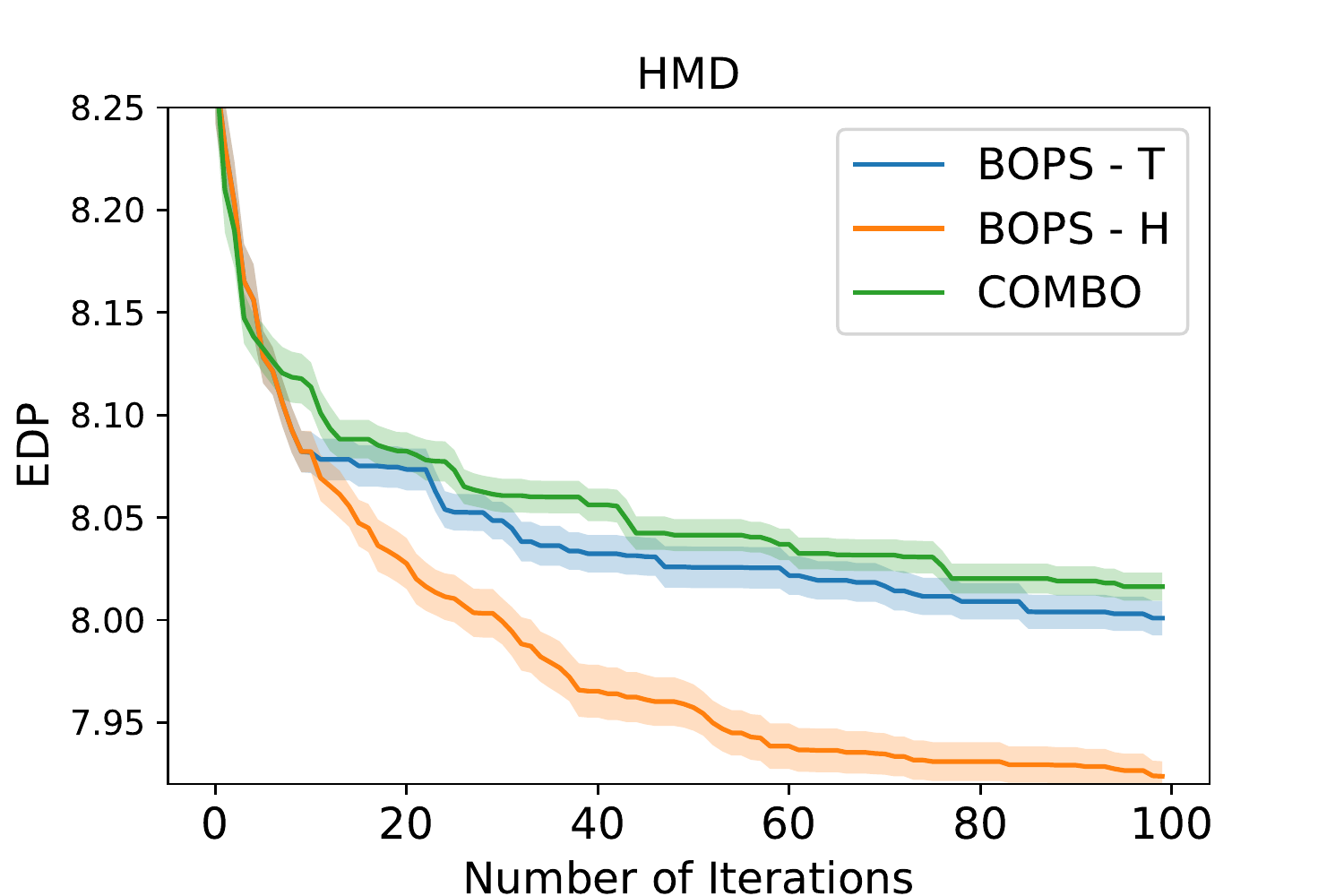}
\label{fig:hmd}
}
\caption{Results comparing BOPS-T, BOPS-H, and COMBO (best objective function value vs. number of BO iterations) on both synthetic and real-world benchmarks: (Top row) QAP, TSP, CP; and (Bottom row) FP1, FP2, and HMD.} 
\label{fig:bo_results}
\end{figure*}

In this section, we describe the benchmarks and experimental setup followed by results and discussion.

\subsection{Benchmarks}

We employ diverse and challenging benchmarks for black-box optimization over permutations for our experiments. We have the following two synthetic benchmarks.


\noindent {\bf 1) Quadratic assignment problem (QAP).} QAPLIB \cite{burkard1997qaplib} is a popular library that contains multiple QAP instances. Each QAP instance contains a cost matrix ($A$) and distance matrix ($B$) sized $n \times n$, where $n$ is the number of input dimensions. 
The goal is to find the best permutation that minimizes the quadratic assignment objective $Tr(A  P B P^T)$, where $P$ is an $n \times n$ permutation matrix. We use input space with $n = 15$ dimensions in our experiments.


\noindent {\bf 2) Traveling salesman problem (TSP).}
TSP problems are derived from low-dimensional variants of the printed circuit board (PCB) problems from the TSPLIB library \cite{tsplib}. The overall goal is to find the route of drilling holes in the PCB that minimizes the time taken to complete the job. 
We use input space with $d=10$ dimensions from the data provided in the library.


We perform experiments on three important real-world applications 
from the domain of computer-aided design of integrated circuits (ICs). These applications are characterized by permutations over functional blocks at different levels of granularity that arise in different stages of design and optimization of ICs. Importantly, even tiny improvements in solution has huge impact (e.g., improved performance over the lifespan of the IC or reduced cost for manufacturing large samples of the same IC). A big challenge in the combinatorial BO literature is the availability of challenging real-world problems to evaluate new approaches. Hence, we provide our three real-world benchmarks as a new resource to allow rapid development of the field.


\noindent {\bf 3) Floor planning (FP).} We are given $k$ rectangular blocks with varying width and height, where each block represents a functional module performing certain task. Each placement of the given blocks is called a {\em floor-plan}. Our goal is to find the floor plan that minimizes the manufacturing cost per chip.  We use two variants of this benchmark with 10 blocks (FP1 and FP2) that differ in the functionality of the blocks.


\noindent {\bf 4) Cell placement (CP).} We are given 10 rectangular cells with same height and a netlist that contains the connection information among the cells. The goal is to place the 10 rectangular cells for optimizing the performance of the circuit. Intuitively, shorter nets have shorter delays, so placements with shorter wire-length will result in higher performance.


\noindent {\bf 5) Heterogeneous manycore design (HMD).} This is a manycore architecture optimization problem from the rodinia benchmark \cite{rodinia-che}. 
We are given 16 cores of three types: 2 CPUs, 10 GPUs, and 4 memory units. They are connected by a mesh network (each core is connected to its four neighboring cores) to facilitate data transfer. The goal is to place the given 16 cores to optimize the energy delay product (EDP) objective that captures both latency and energy, two key attributes 
of a manycore chip.

\begin{figure*}[h!]
\centering
\subfloat{
\includegraphics[width=0.30\textwidth]{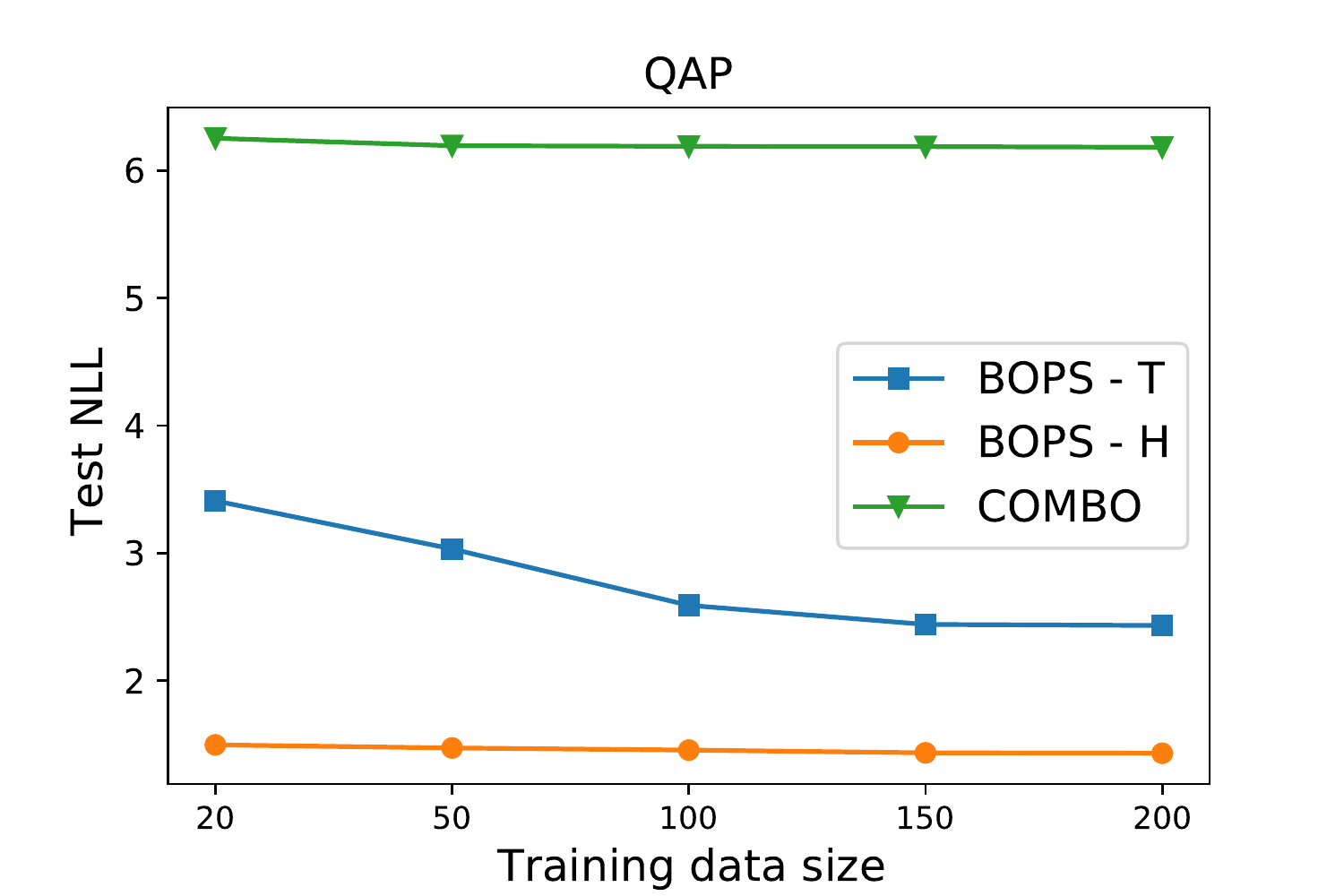}
\label{fig:qap}
}
\subfloat{
\includegraphics[width=0.30\textwidth]{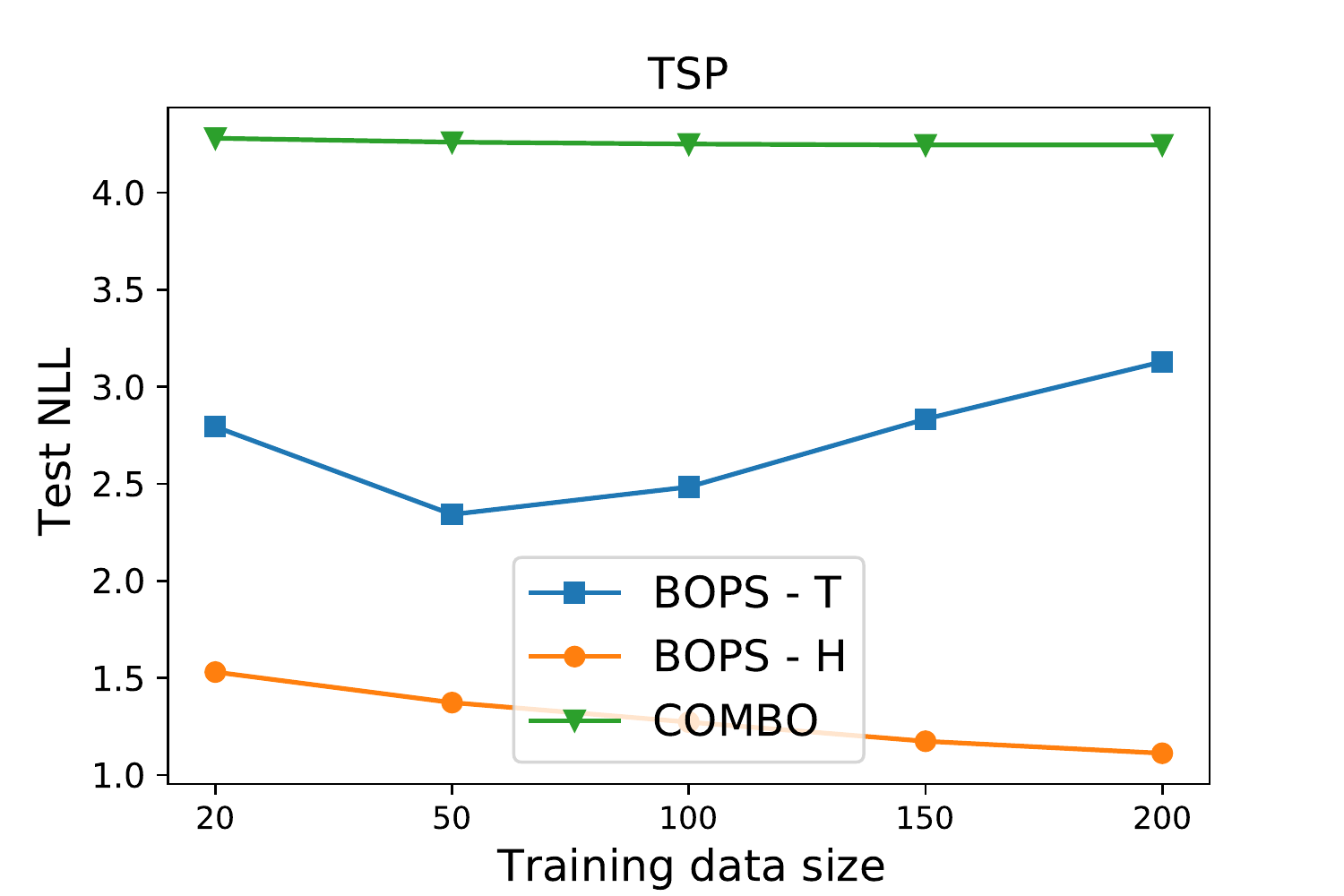}
\label{fig:tsp}
}
\subfloat{
\includegraphics[width=0.30\textwidth]{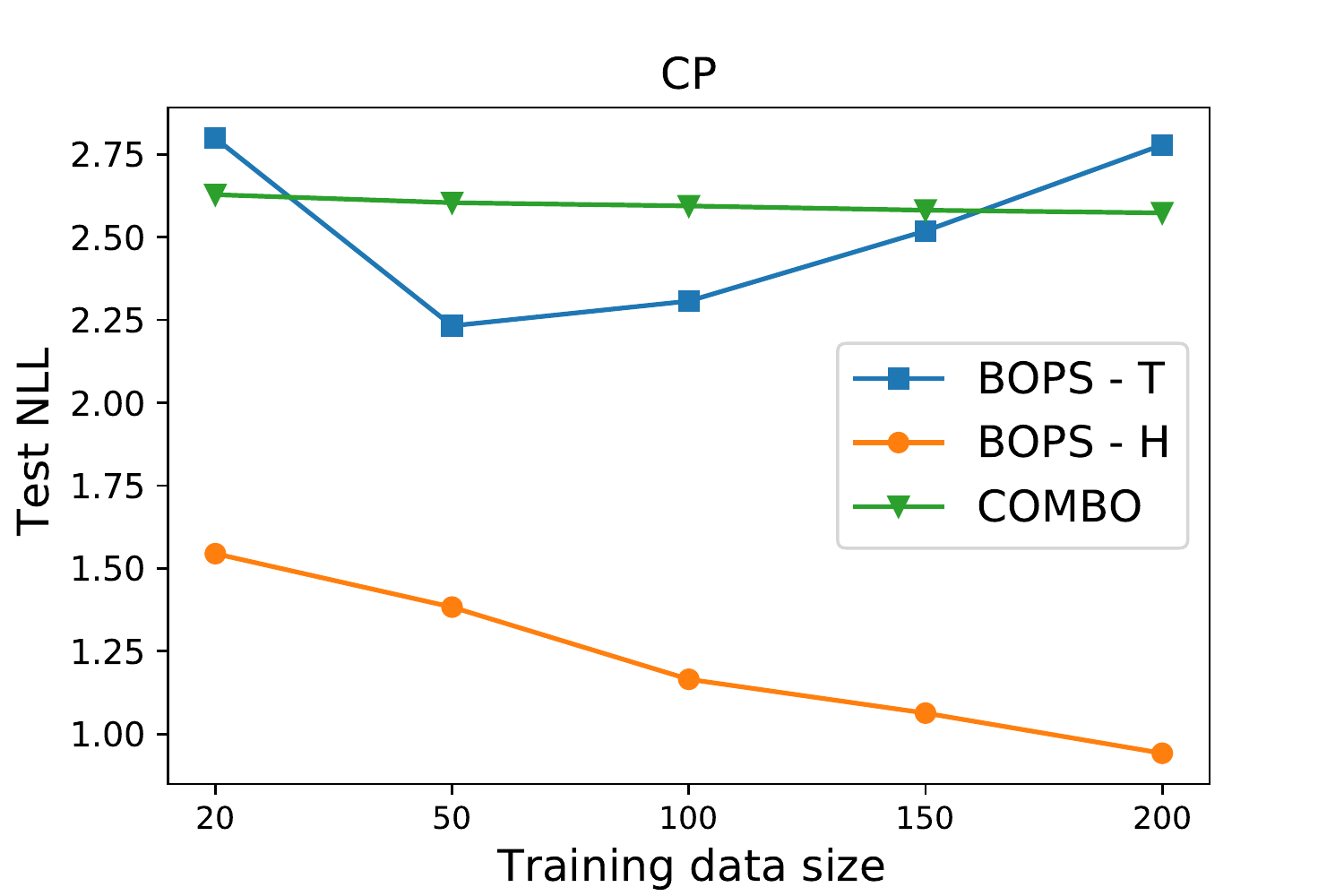}
\label{fig:pl}
} 
\quad
\subfloat{
\includegraphics[width=0.30\textwidth]{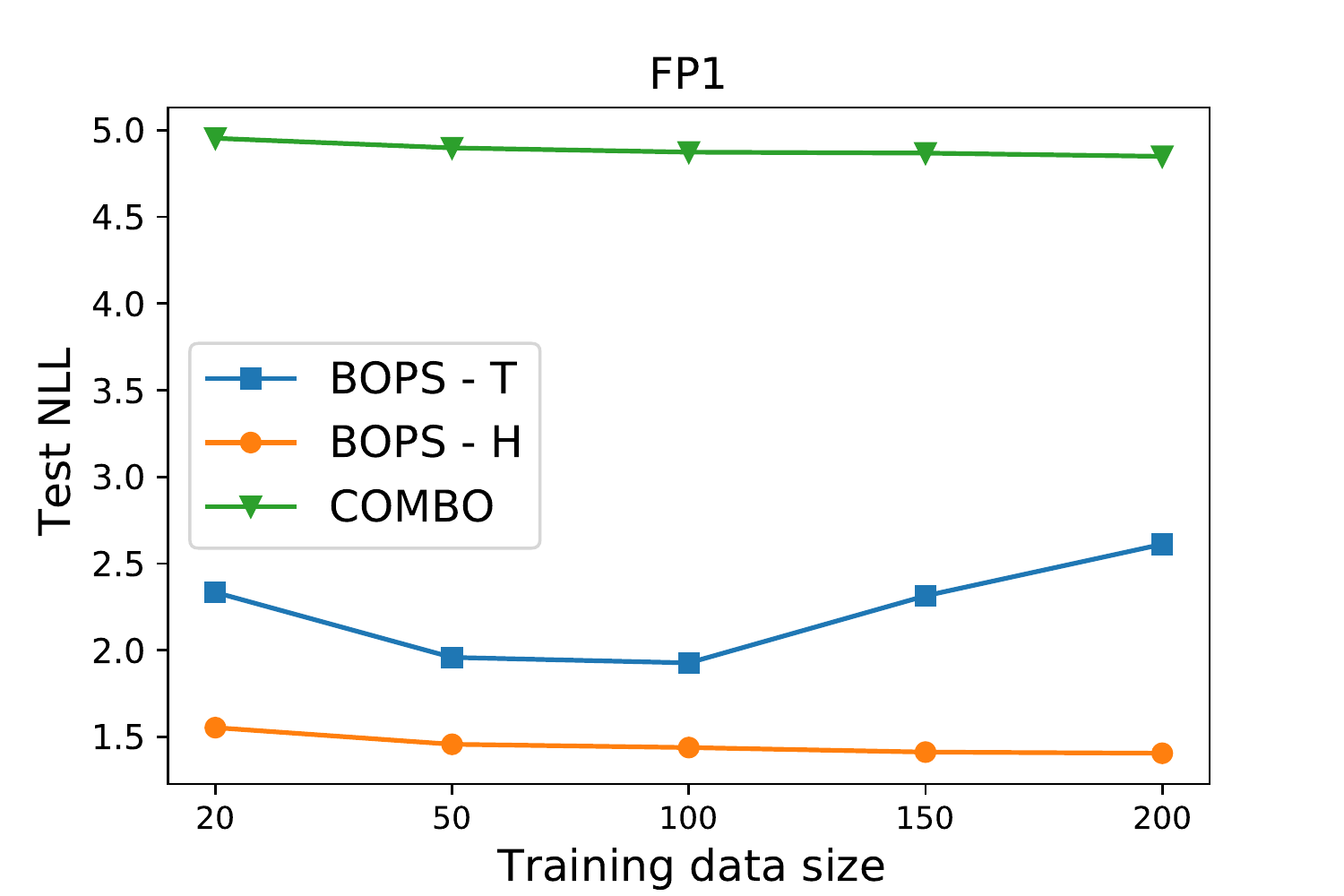}
\label{fig:fp_1}
}
\subfloat{
\includegraphics[width=0.30\textwidth]{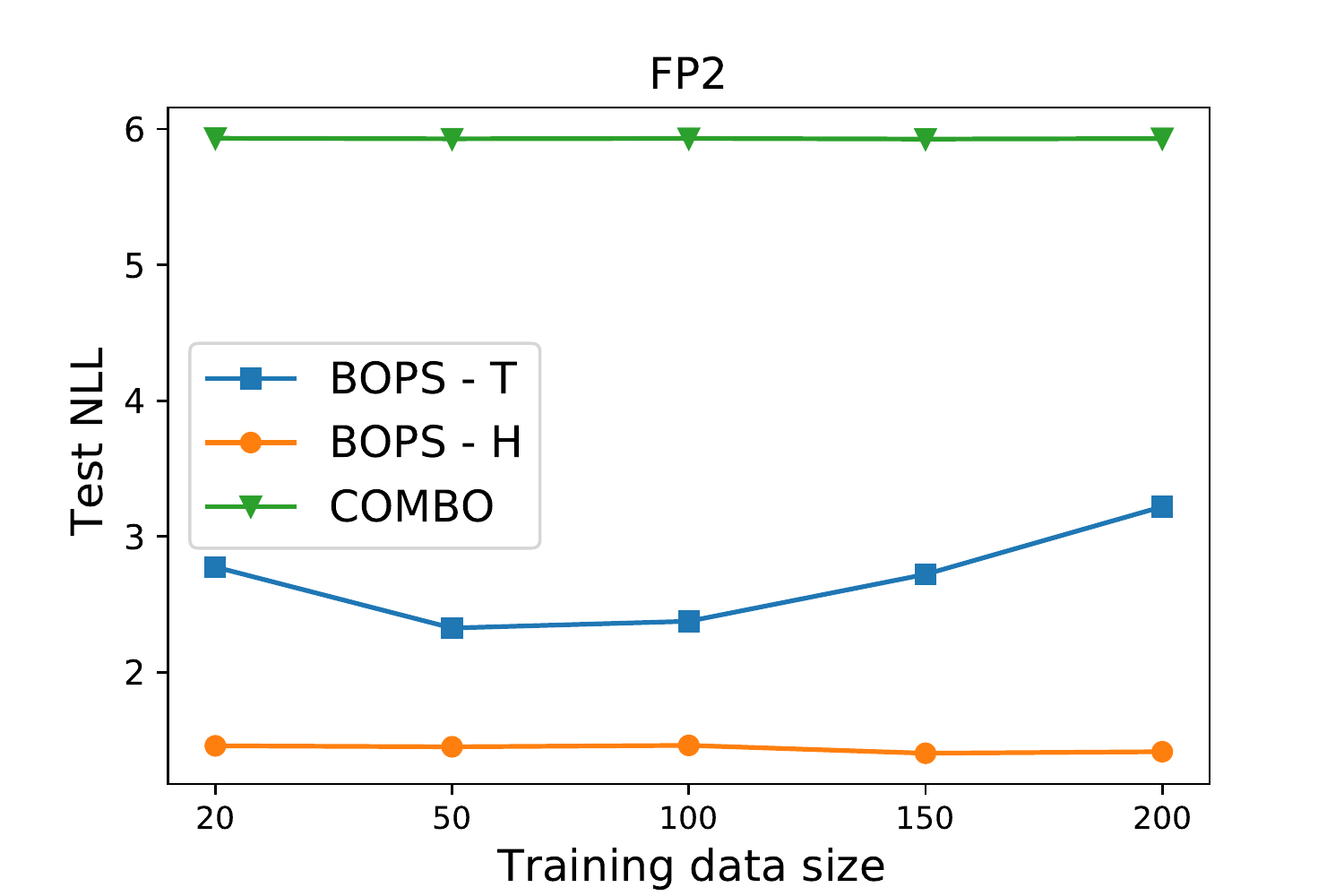}
\label{fig:fp_2}
}
\caption{Results comparing the three surrogate models of BOPS-T, BOPS-H and COMBO on negative log-likelihood (NLL) metric computed on a test set on five benchmarks: (Top row) QAP, TSP, CP; and  (Bottom row) FP1, FP2.} 
\label{fig:nll_results}
\end{figure*}

\subsection{Experimental Setup}


\noindent {\bf Configuration of algorithms.} We compare our proposed BOPS-T and BOPS-H algorithms with the state-of-the-art combinatorial BO algorithm COMBO \cite{COMBO}. COMBO employs a diffusion kernel based GP surrogate model and optimizes expected improvement acquisition function using local search with restarts to select inputs for evaluation. Each local search step considers all neighbors of the current structure in the combinatorial graph (i.e., structures with Hamming distance one). We modify COMBO's local search procedure (\url{https://github.com/QUVA-Lab/COMBO}) to consider only those neighbors which are permutations of the current state thereby helping COMBO to avoid searching a large combinatorial space with huge number of invalid structures (non-permutations). 

We used the  SDP relaxation based QAP solver code from (\url{https://github.com/fsbravo/csdp}) for implementing BOPS-T.  BOPS-H is built using popular GPyTorch \cite{gardner2018gpytorch} and BoTorch \cite{balandat2020botorch} libraries. We used 10  restarts for local search based EI optimization for BOPS-H. 
BOPS-T, BOPS-H, and COMBO are initialized with the same 20 random permutations in each experiment. 


\noindent {\bf Evaluation metric.} We plot the objective function value of the best permutation over different BO iterations. Each experiment is repeated 20 times and we plot the mean of the best objective value plus and minus the standard error.

\subsection{Results and Discussion}

In this section, we present and discuss our experimental results along different dimensions.


Figure \ref{fig:bo_results} shows the results for BO performance (best objective value vs. number of function evaluations / BO iterations) of BOPS-T, BOPS-H, and COMBO on all six benchmarks. Below we discuss these results in detail.


\noindent {\bf BOPS-T vs. BOPS-H.} Recall that BOPS-T and BOPS-H makes varying trade-offs between the complexity of statistical model and tractability of acquisition function optimization: BOPS-T uses simple model and tractable search; and BOPS-H employs complex model and heuristic search. From Figure \ref{fig:bo_results}, we can observe that BOPS-H performs significantly better than BOPS-T on all six benchmarks. 


\noindent {\bf BOPS vs. COMBO.} From the results shown in Figure \ref{fig:bo_results}, we make the following observations: 1) BOPS-H performs significantly better than both BOPS-T and COMBO on all six benchmarks; and 2) BOPS-T is comparable or slightly better than COMBO on all benchmarks except TSP and CP. 


We hypothesize that the performance of different BO algorithms, namely, BOPS-H, BOPS-T, and COMBO is proportional to the quality of their surrogate models in terms of making predictions on unknown permutations and their uncertainty quantification ability. To verify this hypothesis, we compare the three surrogate models quantitatively in terms of their performance on the log-likelihood metric.

\noindent {\bf Comparison of surrogate models.} We compare the three surrogate models on the log-likelihood metric \cite{murphy2012machine} because it captures both the prediction and uncertainty quantification of a model which are essential for the effectiveness of BO. We plot the negative log-likelihood (NLL) of the three surrogate models on a testing set of 50 instances as a function of the increasing size of training data. Each experiment is replicated with 10 different training sets and each method is evaluated using the median of the NLL metric on 10 different test sets of 50 permutations each. Figure \ref{fig:nll_results} shows the results on all benchmarks except HMD. We do not show results on HMD since each function evaluation is much more expensive when compared to all other benchmarks,  and we are generating multiple replications of the training and testing sets (10 $\times$ 10 = 100 runs). We make the following observations from Figure \ref{fig:nll_results}: 1) BOPS-H shows the best performance among the three surrogate models; 2) BOPS-T does better than COMBO on all benchmarks other than cell-placement. Since both COMBO and BOPS-H employ the same acquisition function (EI) and optimizer (local search), it is evident that the gains in the BO performance comes from the superior surrogate model of BOPS-H.

%% file: files/appendix.tex
\section{Additional Experiments}
\begin{figure}[h!]
\centering
\subfloat[QAP (15 dimensions)]{
\includegraphics[width=0.30\textwidth]{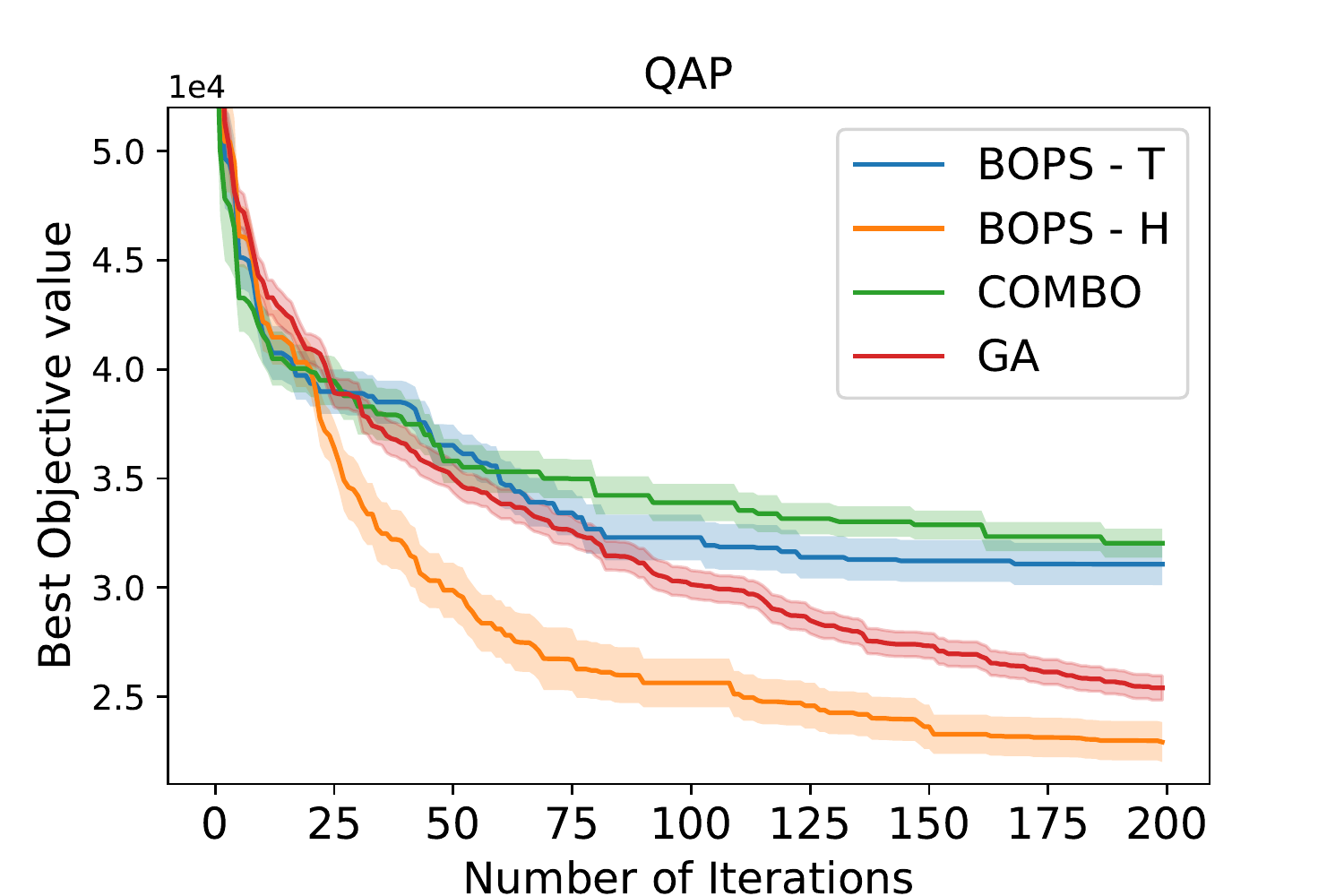}
\label{fig:qap_15}
}
\subfloat[TSP (10 dimensions)]{
\includegraphics[width=0.30\textwidth]{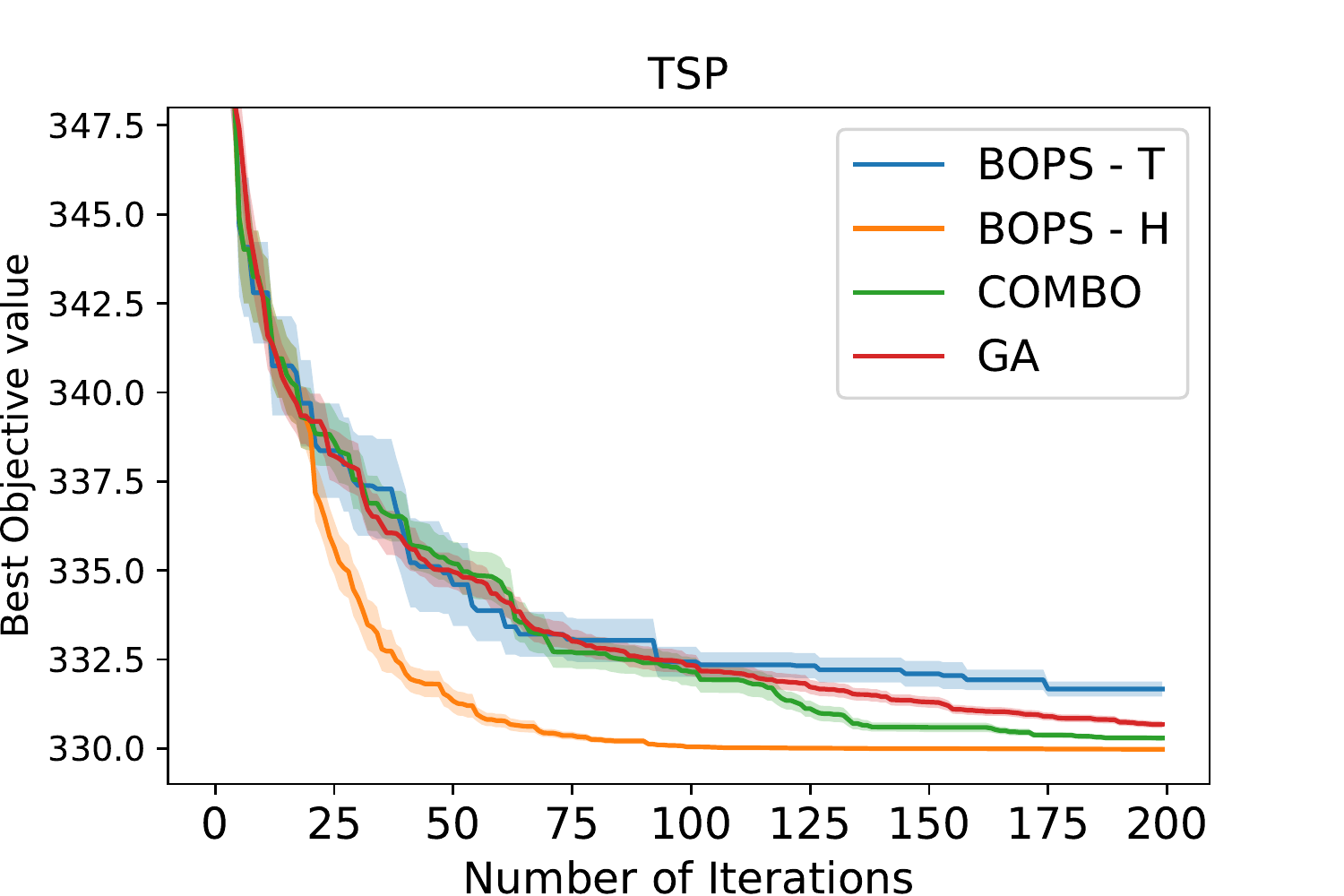}
\label{fig:tsp_10}
} 
\quad
\subfloat[Cell Placement]{
\includegraphics[width=0.30\textwidth]{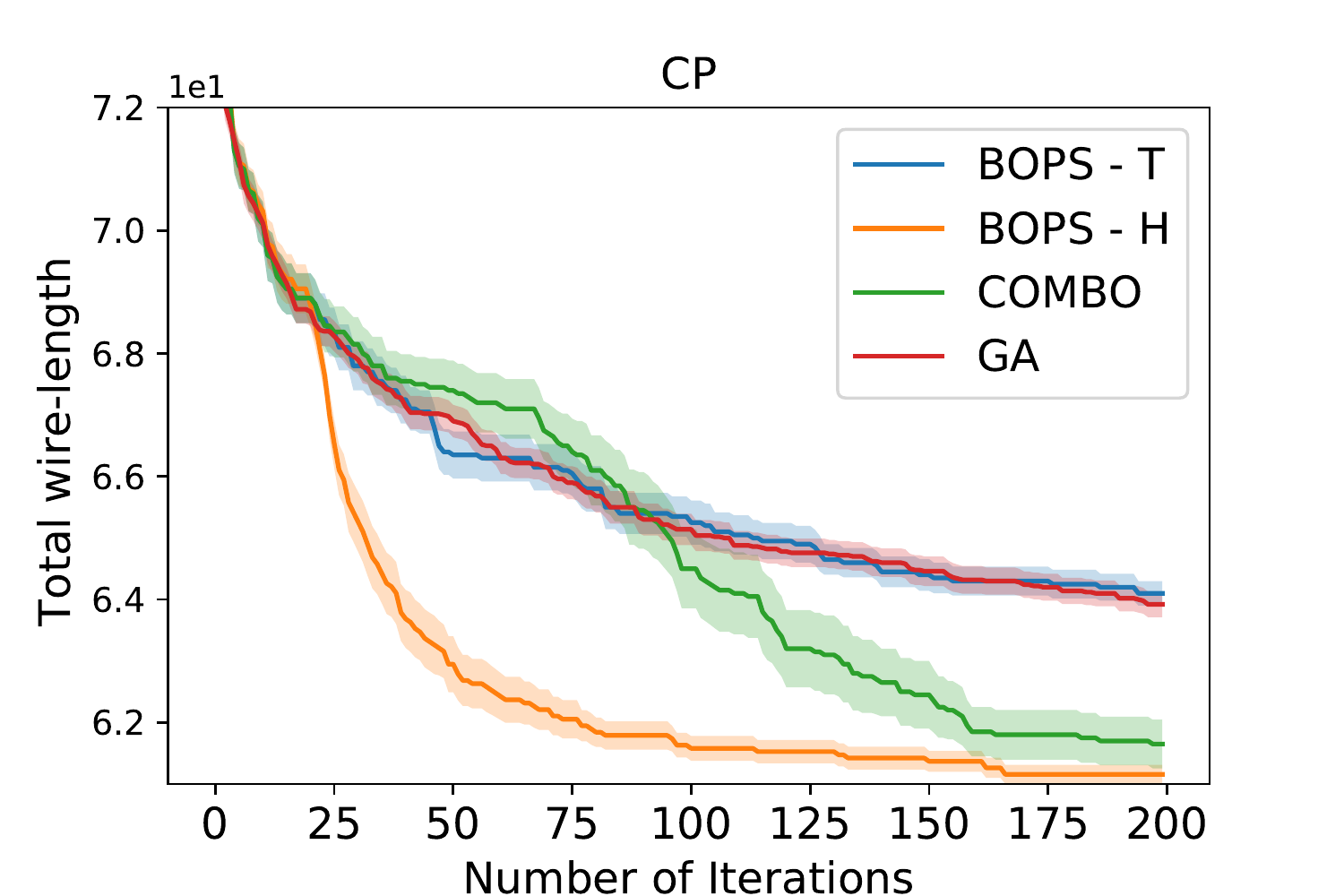}
\label{fig:cp}
}
\subfloat[Floorplanning 1]{
\includegraphics[width=0.30\textwidth]{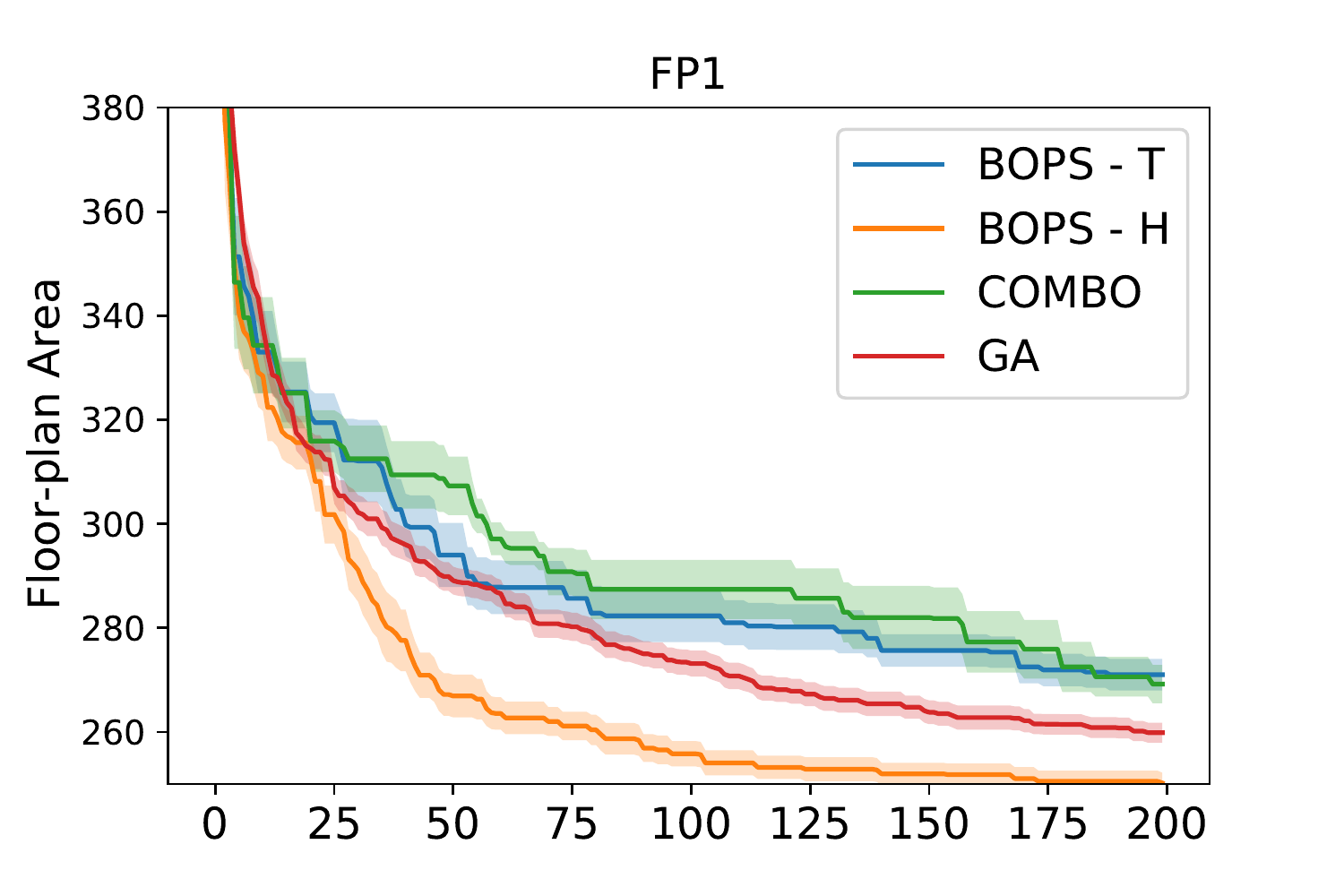}
\label{fig:fp_1}
}\quad
\subfloat[Floorplanning 2]{
\includegraphics[width=0.30\textwidth]{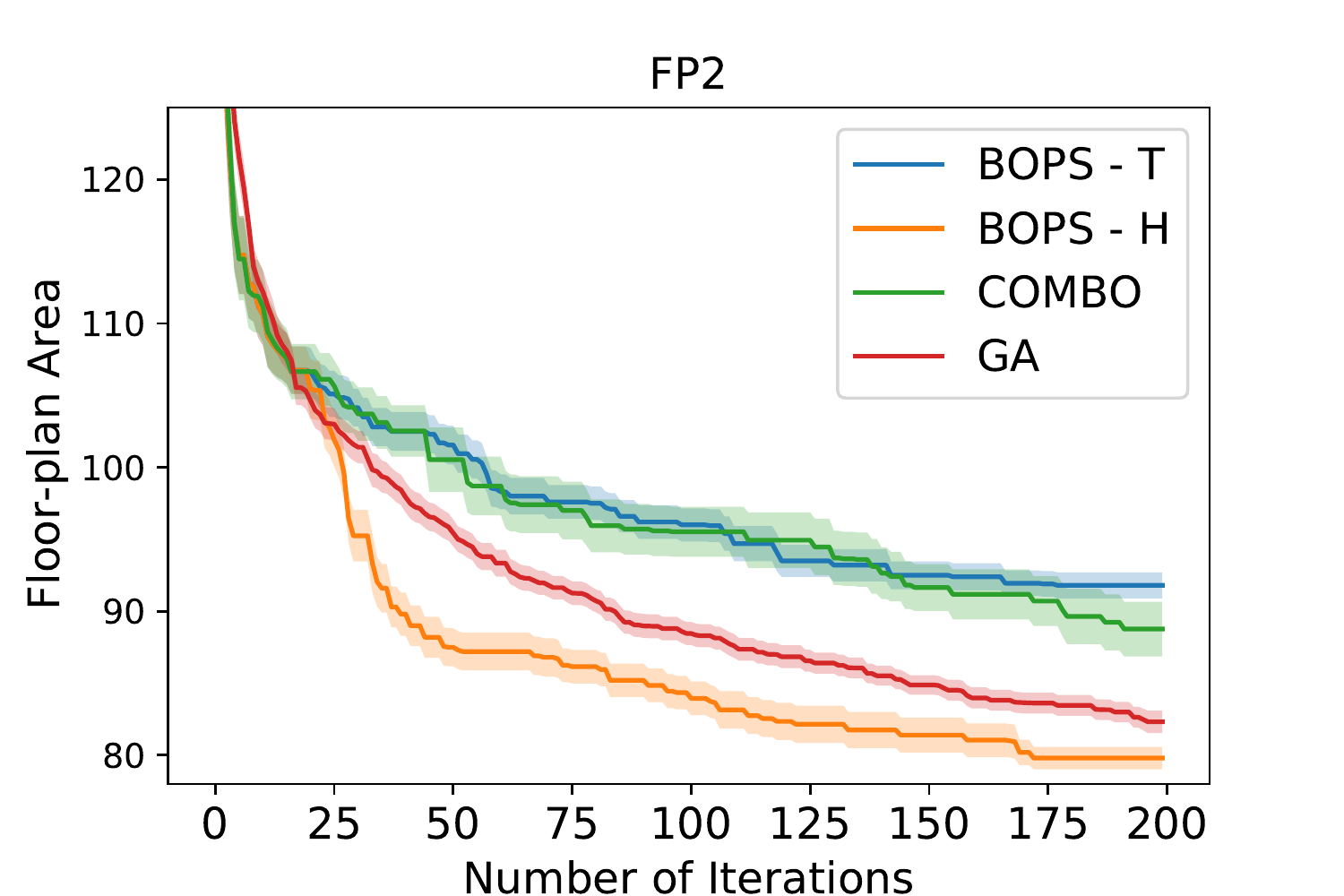}
\label{fig:fp_2}
}
\caption{Results comparing BOPS-T, BOPS-H, and COMBO and Genetic Algorithm (GA) (best objective function value vs. number of BO iterations) on both synthetic and real-world benchmarks:} 
\label{fig:ga_results}
\end{figure}

Genetic algorithms \cite{davis1991handbook} are a commonly employed technique for optimizing discrete structures. They have also been used for optimizing acquisition functions \cite{string_bo} over discrete spaces in Bayesian optimization. We present additional results comparing our proposed algorithms with genetic algorithms in Figure \ref{fig:ga_results}. We used a genetic algorithm with permutation based operators as implemented in Platypus \url{https://github.com/Project-Platypus/Platypus} library. We tried few different choices for the two hyper-parameters (population size and offspring size) and picked population size to be 20 and offspring size to 10. Since we are in a small data setting, large values for these hyper-parameters will not be practically useful. As seen in the figure, BOPS-H still performs the best outperforming even genetic algorithms demonstrating the importance incorporating a model-guided search like Bayesian optimization for finding the best structure.